\documentclass[runningheads]{llncs}
\usepackage[T1]{fontenc}
\usepackage{graphicx}
\usepackage{float}
\usepackage[caption=false]{subfig}
\usepackage{multirow}
\usepackage{booktabs}
\usepackage{graphicx}
 \usepackage{threeparttable}

\usepackage[pagebackref=true,breaklinks=true,colorlinks,bookmarks=false]{hyperref}
\usepackage{color}

\begin{document}
\bibliographystyle{plain}

\title{ROFusion: Efficient Object Detection using Hybrid Point-wise Radar-Optical Fusion}
\titlerunning{ROFusion}

\author{Liu Liu \and
Shuaifeng Zhi\thanks{Shuaifeng Zhi is the corresponding author.} \and
Zhenhua Du \and
Li Liu \and
Xinyu Zhang \and
Kai Huo \and
Weidong Jiang
}
\authorrunning{L. Liu et al.}
\institute{College of Electronic Science, National University of Defense Technology
\\
\email{liuliucn@outlook.com, zhishuaifeng@outlook.com}
}
\maketitle              %
\begin{abstract}

Radars, due to their robustness to adverse weather conditions and ability to measure object motions, have served in autonomous driving and intelligent agents for years. 
However, Radar-based perception suffers from its unintuitive sensing data, which lack of semantic and structural information of scenes. To tackle this problem, camera and Radar sensor fusion has been investigated as a trending strategy with low cost, high reliability and strong maintenance. While most recent works explore how to explore Radar point clouds and images, rich contextual information within Radar observation are discarded. In this paper, we propose a hybrid point-wise Radar-Optical fusion approach for object detection in autonomous driving scenarios. The framework benefits from dense contextual information from both the range-doppler spectrum and images which are integrated to learn a multi-modal feature representation. Furthermore, we propose a novel local coordinate formulation, tackling the object detection task in an object-centric coordinate. Extensive results show that with the information gained from optical images, we could achieve leading performance in object detection (97.69\% recall) compared to recent state-of-the-art methods FFT-RadNet \cite{rebut:2022raw} (82.86\% recall). Ablation studies verify the key design choices and practicability of our approach given machine generated imperfect detections. The code will be available at \url{https://github.com/LiuLiu-55/ROFusion}.

\end{abstract}

\section{Introduction}

Autonomous driving and Advanced Driver Assistance Systems (ADAS) often rely on different types of sensors to acquire a reliable perception. Mainstream sensors equipped in automotive vehicles are camera, Lidar and Radar, which are fused together thanks to their unique working mechanism and specialties.
Existing mainstream multi-sensor fusion strategy uses camera and Lidar sensors for 3D object detection \cite{chen：2017multilidar,vora:2020pointpainting}. Mainly because Lidar owns a high angular resolution and range detection accuracy in a way of dense point clouds, and is complementary to camera images which are rich in contextual and semantic information of scenes. However, both camera and Lidar suffer from huge performance degradation in adverse weather conditions, which is a crucial requirement for long-term stable autonomous driving. 

Radars are active sensors that measure the environment from reflected electromagnetic waves.
Compared to Lidar, Radar has a robust capacity in severe weather conditions and can detect objects and obstacles within 250m with their distances and velocities. Furthermore, its low deployment cost makes Radar a requisite sensor in assistance systems. 
Radar data have developed different types of representations, including Radar occupancy grid maps, micro-Doppler signature, dense Range-Doppler-Azimuth (RAD) tensors and point clouds, with various processing costs and representational capacity. 

Despite Radar's advantages in stable and long-term scene perception, there have been few investigations on fusing Radar with other sensors in this task. This is partly caused by its entirely different imaging mechanism in contrast to cameras and Lidars, leading to extremely sparse point clouds or intuitive dense RAD spectrum, and
relatively low elevation angular resolution as well.
Fortunately, this problem has been partly solved with the development of the 4D imaging Radar, with a high angular resolution of about 1° in both azimuth and elevation. Some recent works also tried to conduct image-Radar fusion to alleviate the high sparsity of Radar point clouds \cite{nabati:2021centerfusion,kim:2022craft,hwang:2022cramnet}.

Motivated by the above-mentioned challenges, we propose ROFusion, a hybrid point-wise approach to fuse Radar and camera data.
Different from previous work in Radar-optical fusion, we seek to fuse dense contextual features from both modalities. We first acquire Radar and camera features respectively from single-modality extractors \cite{rebut:2022raw,he:2016res18}, and then use image-Radar association and hybrid point-based fusion strategy to merge cross-modality features at multiple hierarchies. Finally, a local coordinate formulation is proposed to decompose our tasks into classification and regression in an object-centric manner. Our method achieves a new state-of-the-art performance in both easy and hard cases of public RADIal dataset  \cite{rebut:2022raw}.

To summarize, our contributions are as follows:
\begin{itemize}
    \item We propose a hybrid point-wise fusion strategy to effectively associate dense Radar and image features.
    \item We propose a local coordinate formulation that simplifies object detection by classification and regression sub-tasks in an object-centric manner.
    \item We conduct extensive experiments on the RADIal \cite{rebut:2022raw} benchmark and achieve a new state-of-the-art detection performance, with a significant boost over Radar-based baseline.
\end{itemize}

\section{Related work\label{sec:related works}}

\subsection{Point-based Methods}

PointNet \cite{qi:2017pointnet} designs a novel type of neural network that directly consumes the point cloud, which makes point-based detection methods process. For Radar point clouds, sparse structures take a challenge to object detection. One strategy \cite{dreher:2020radar} is to accumulate radar points into a dense occupancy grid mapping (OGM). For lightweight demand, \cite{scheiner:2021object} utilize novel point structure \cite{qi:2017pointnet++}. With the sparsity issue, \cite{liu:2022deep} observes that a global message could enhance perception performance. 

\subsection{Camera-Radar Fusion Methods}

Complementary information gives the opportunity for sensor fusion between the camera and Radar. Radar extracts the distance and velocity of objects, while semantic information is captured by cameras. 
There are normally three fusion levels between Radar and camera: early level, feature level, and late level.
Radars are often used to generate the region of interest (ROI) for early-level fusion. Then, the predicted region is processed as an auxiliary refining optical task \cite{gaisser:2017road,guo:2018pedestrian}, which is computationally expensive. The decision level contrary utilizes two sources independently detect, proposing a strategy \cite{cesic:2016radar,zhong：2018camera} defining whether one of the sensors failed. With different probability spaces, late-level fusion could not efficiently exert the capability of two sensors. 

A naive approach is fusing Radar and camera in a latent feature space where the key point is Radar-camera association. CramNet \cite{hwang:2022cramnet} applies a dynamic voxelization fusing Radar and camera features, projecting each camera pixel with a 3D ray to rectify its location, which makes a robust performance for sensor failure.
In \cite{nabati:2021centerfusion}, authors propose a frustum association that fully exploits Radar vertical information. 
CRAFT \cite{kim:2022craft} also associates Radar and image, but implements them in a polar coordinate to handle the discrepancy between the coordinate system and spatial properties. The feature maps are then fused by a consecutive cross-attention strategy.

\begin{figure}[t!]
\includegraphics[width=\textwidth]{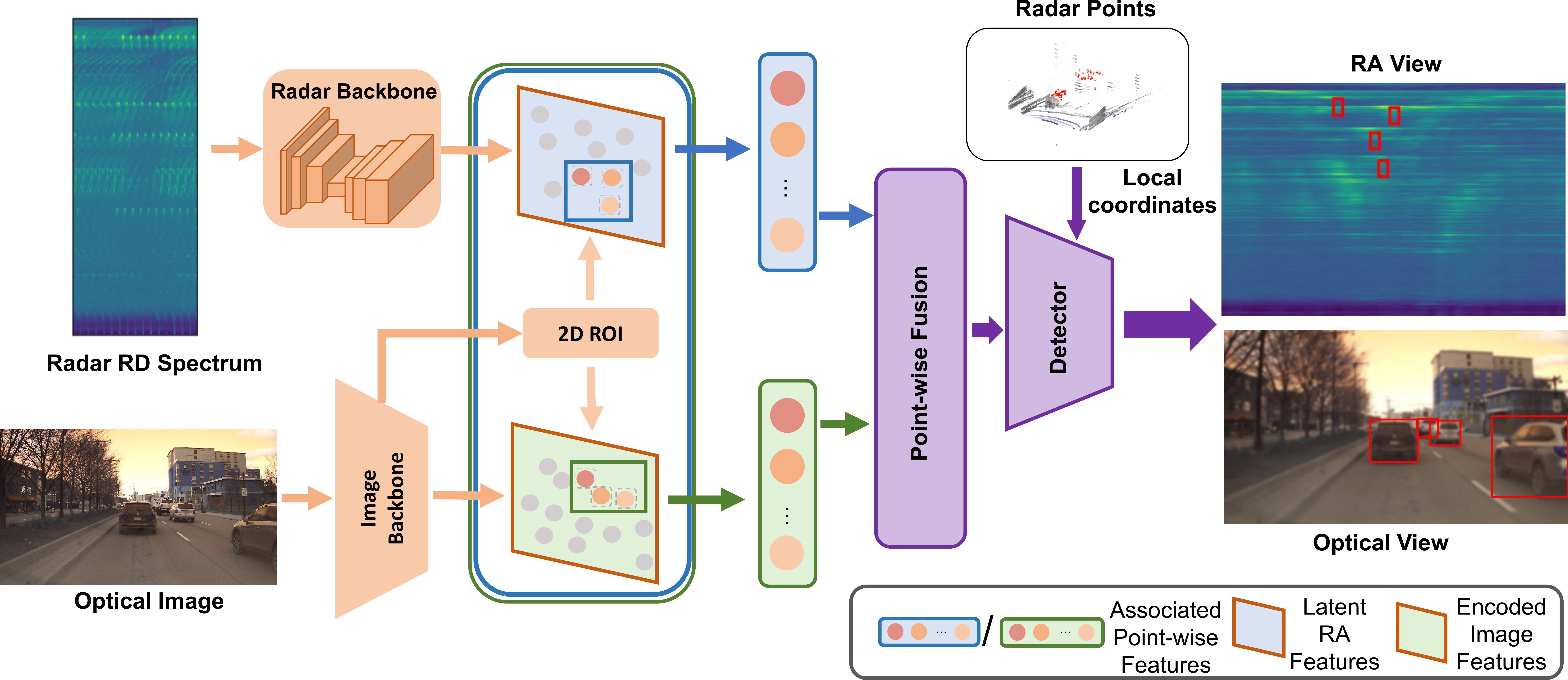}
\caption{ROFusion network architecture. Latent RA feature maps and camera-encoded semantics are first obtained by Radar and image backbone. The image 2D bounding boxes are used to associate the image and Radar feature maps via point cloud candidates. The point-wise module is next used for feature-level fusion, proposing a hybrid point feature to produce final Radar detection formulated in object-centric local coordinates.} \label{network}
\vspace{-0.5cm}
\end{figure}

\section{Method\label{sec:method}}

In this section, we present architectural details of our method ROFusion as well as key design choices enabling accurate object detection with a hybrid point-wise optical-Radar fusion. 
An overall architecture is provided in Figure~\ref{network}. We first take Radar RD spectrums with corresponding images as our network inputs and extract their dense features. Radar points filtered by prior information such as image detection bounding boxes are then adopted as anchors to associate Radar and image features. Furthermore, a hybrid point-wise fusion complements the surrounding semantics of targets to produce new point features. A detection header finally predicts object locations in per-object local coordinates. 

To summarize, our pipeline consists of three main modules including: (1) a dense feature extraction module from both RGB images and corresponding high-definition RD tensors to acquire contextual information of scenes (Section \ref{sec:feat_extract}); (2) a hybrid point-based fusion module to associate dense Radar embedding of scattering points with image features (Section \ref{sec:fusion}); (3) a local coordinate module formulating object detection task in an object-centric manner (Section \ref{sec:loc_coord}). We finally show the training configurations of our method in Section \ref{sec:obj_detec}.

\subsection{Dense Feature Extraction \label{sec:feat_extract}}

In order to acquire rich contextual information about objects within 3D scenes, we leverage dense convolutional neural networks (CNNs) to extract dense feature embedding of both Range-Doppler (RD) Radar maps and camera image observations.

\paragraph{Radar Feature Extractor}
Radar-based scene understanding from its Range-Doppler (RD) map has recently gained attention as it contains all information on range, azimuth and elevation.
In addition, the RD map owns less computational acquisition costs and is a dense representation compared to Range-Doppler-Azimuth (RAD) tensors and sparse point clouds, respectively.
We propose to use a dense CNN model as our Radar backbone module, inspired by FFT-RadNet \cite{rebut:2022raw}. Specifically, it aims to learn a multi-scale dense representation of Range-Azimuth (RA) maps from their input RD counterparts, with a tailored MIMO pre-encoder \cite{rebut:2022raw}.
In this manner, we seek to learn a dense feature embedding of RA maps as they are closely related to downstream vehicle detection tasks.

\paragraph{Image Feature Extractor}
To enrich radar features with optical image features, we encode the corresponding RGB image into a dense feature embedding with a vision CNN model. To reduce computational overhead, we simply use an ImageNet \cite{russakovsky:2015imagenet} pre-trained ResNet-18 model \cite{he:2016res18} and keep the weights intact during training. Please note that our image backbone module could be replaced by stronger vision models such as ResNet-152 \cite{he:2016res18} and vision Transformers \cite{vaswani:2017attention}, depending on the computation budgets.

\subsection{Point Fusion\label{sec:fusion}}

\paragraph{Image-Radar-Association}
In this section, we explain how to establish a cross-modality association of target objects with provided sensor calibration information and prior optical detection results.

As dense features of RA maps and images are difficult to conduct dense alignment due to their different imaging mechanism, we rely on Radar point clouds to bridge them at point-level. Specifically, we represent each Radar point as a 3D point $p=(r, a,d,u,v,x,y,z)$, where $(r, a, d)$ and $(x, y, z)$ are its locations within RAD tensors and real word coordinates, respectively. With the intrinsic and extrinsic of the camera model, we transform Radar points into image coordinates as follows:
\begin{equation}
    u = f_x\frac{x'}{z'}-p_x, \qquad
    v = f_y\frac{y'}{z'}-p_y,
\end{equation}
where $(f_x,f_y,p_x,p_y)$ are camera intrinsic parameters, $(x',y',z')$ is 3D position within camera coordinate transformed by camera extrinsic $[R|t]$ and $(x,y,z)$.

\begin{figure} 
[t!]
\captionsetup[subfigure]{labelformat=empty}
	\centering
	\subfloat[\label{asso:a}]{
		\includegraphics[width=0.36\textwidth]{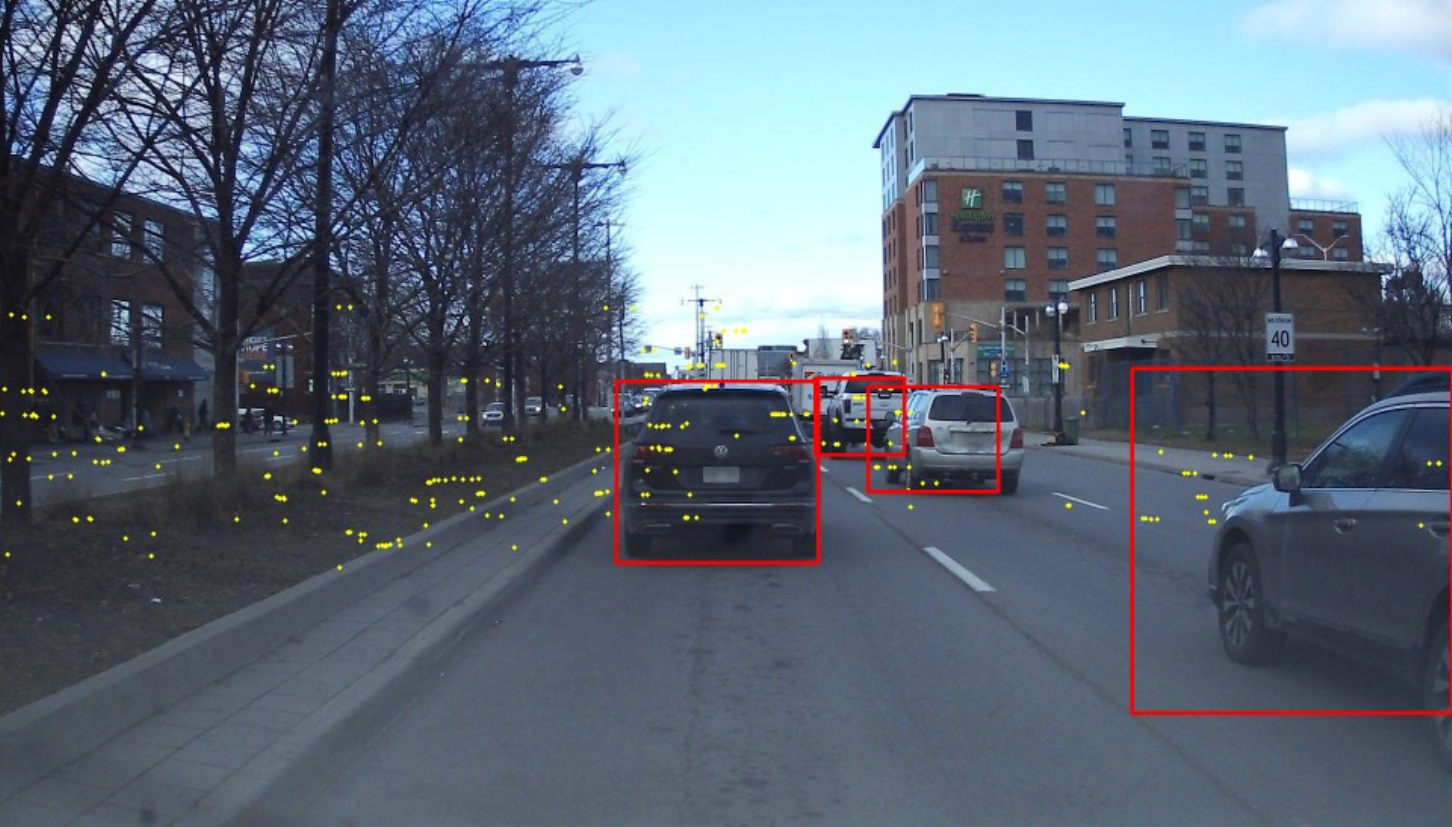}}
	\subfloat[\label{asso:b}]{
		\includegraphics[width=0.26\textwidth]{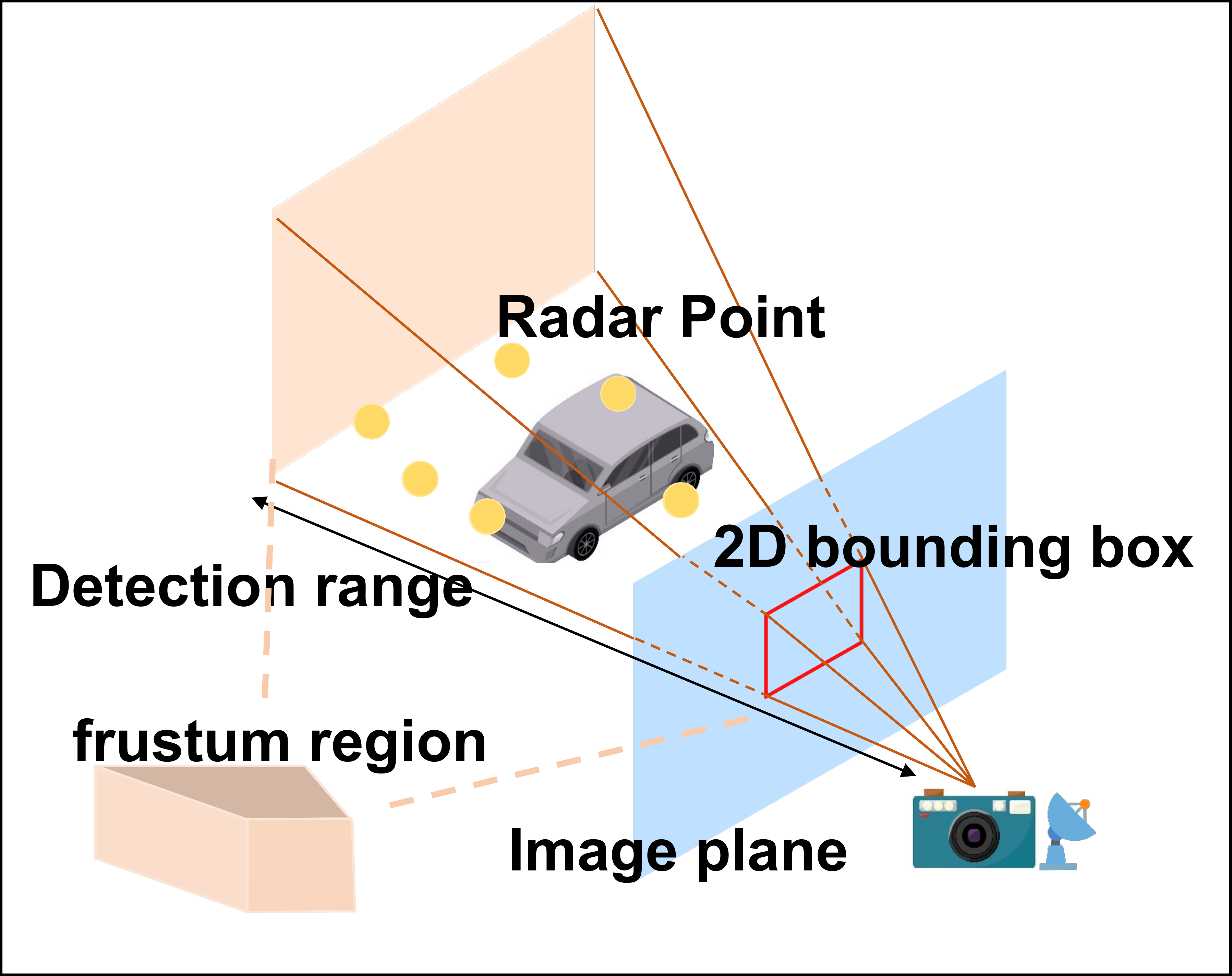}}
	\subfloat[\label{asso:c}]{
		\includegraphics[width=0.31\textwidth]{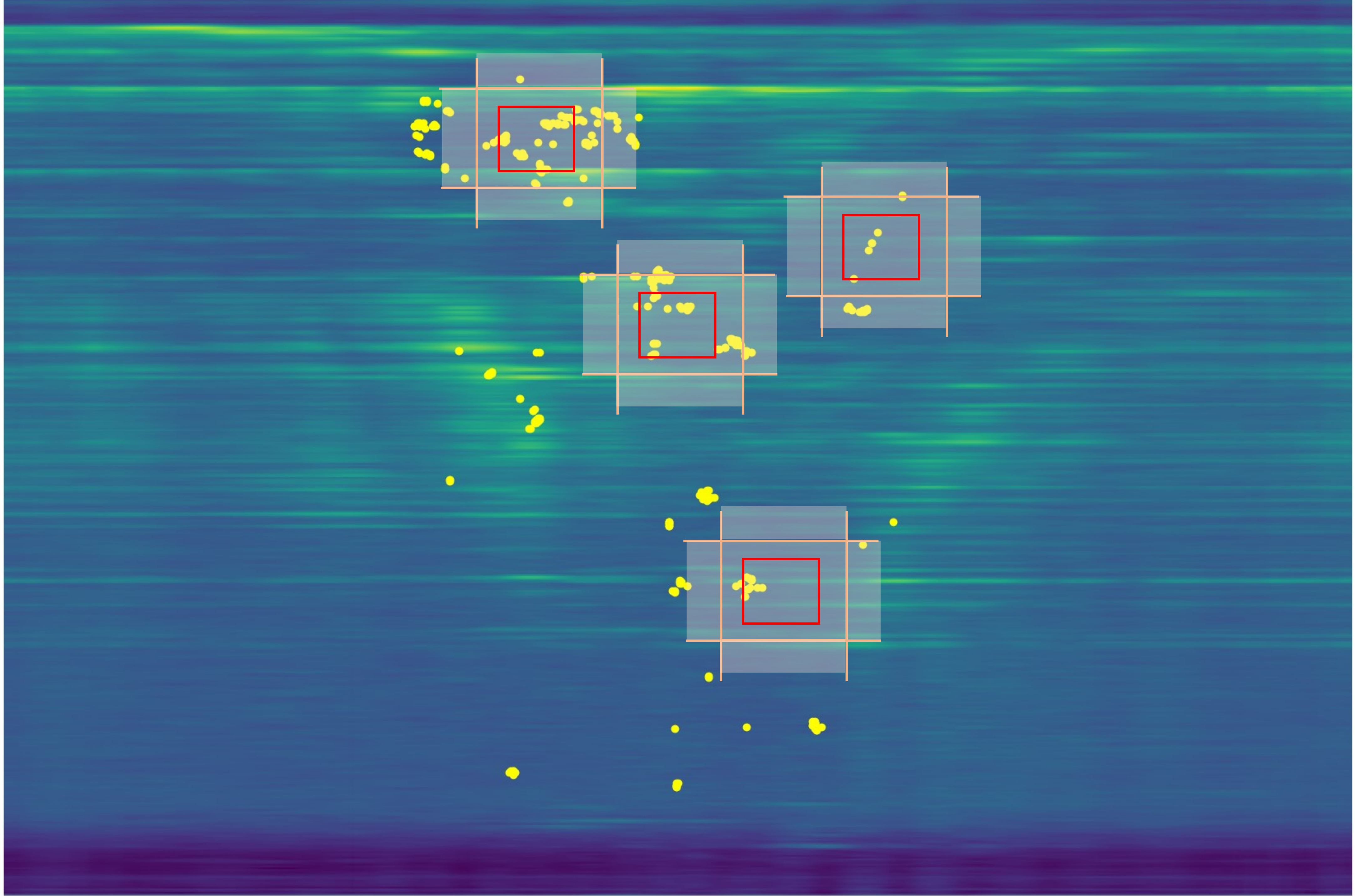}}
  
        \vspace{-2em}
	\caption{Image-Radar Association. 2D Detector using image features provides the azimuth of interest (left), which leads to a frustum region to select candidate object-related point clouds (middle). We filter noise and background points depending on whether their relative radial or angular distance to the object center is beyond a certain threshold or not, as discussed in Section \ref{sec:loc_coord}  (right).}
	\label{association} 
 \vspace{-0.5cm}
\end{figure}

2D object bounding boxes within images are treated as Region of Interest (ROI) filters separating the region of interests out of background and noises, as explained in Figure \ref{association}. The 2D bounding boxes provide strong prior angular information of objects, eliminating the uncertainty caused by Radar sidelobe jamming. At this stage, we treat all points within these 2D ROIs as candidate points for the next fusion stage.
However, these boxes within images cannot cope with range estimation, as points within 3D space in the cone area (middle of Figure \ref{association}) all project within the 2D ROIs.
To address the range inaccuracy, we consider a local coordinate strategy as detailed in Section \ref{sec:loc_coord}. 

\paragraph{Hybrid Point Fusion}
We propose a point-based method that generates per Radar point fused feature from pixel-level RA and image features. Inspired by DenseFusion \cite{wang2019:densefusion}, we implement a variant architecture that fuses semantics and velocity.

\begin{figure}
[t!]
\includegraphics[width=\textwidth]{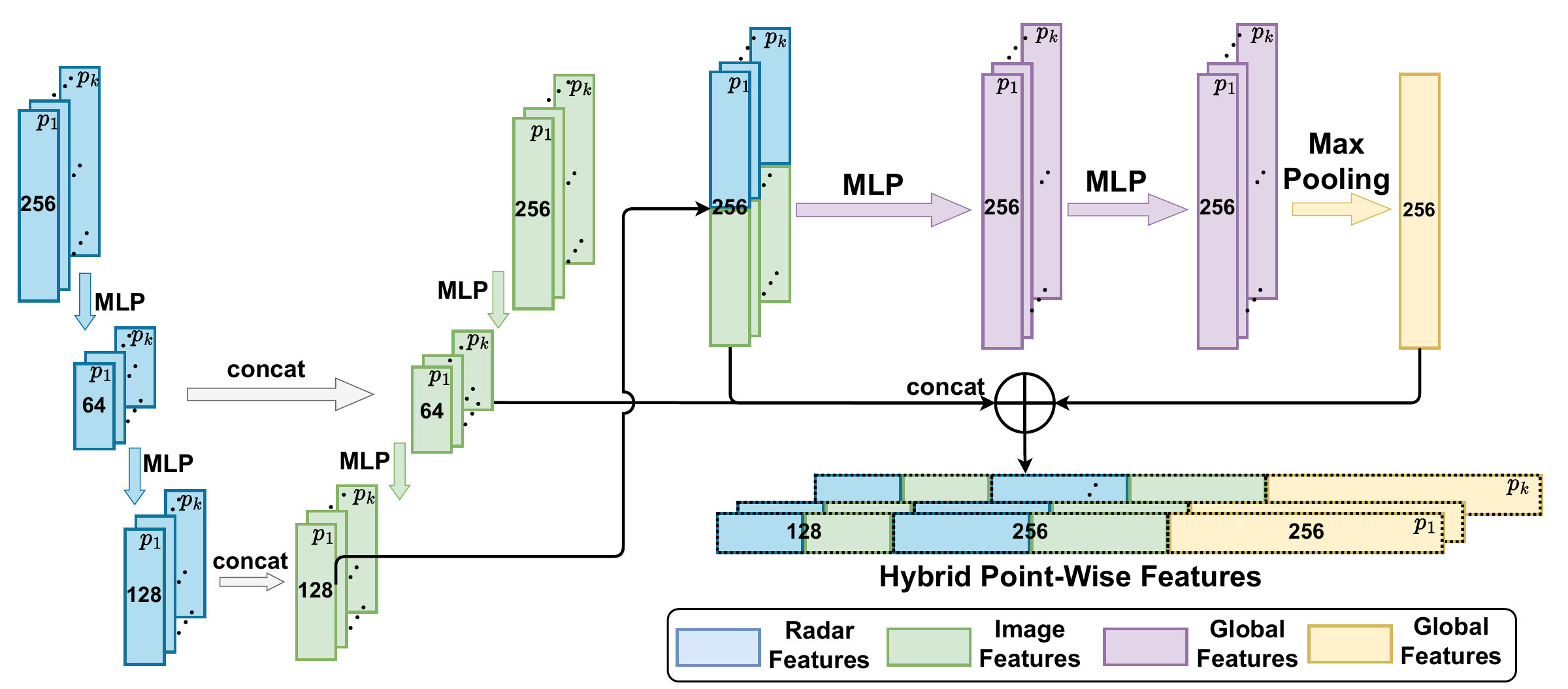}
\caption{Hybrid point fusion architecture. Extracted dense Radar and image features are processed by a series of MLPs and fused at multiple scales. With a per-object global feature from max-pooling across points, we reach a hybrid multi-modal feature with spatial and semantic information by concatenating all fused features of various scales. \label{fusion}}
\vspace{-0.5cm}
\end{figure}

Assume there are $k$ 3D Radar points from the previous association stage, we collect pixel-wise features from encoded RA features $F_R$ and semantic image features $F_I$, respectively. Concretely,
with a set of $k$ point clouds $P=\{p_1,p_1,...,p_k\}$, we extract corresponding per-pixel features $F_R=\{F_r^{p_1},F_r^{p_2},...,F_r^{p_k}\}$ and RA features $F_I=\{F_i^{p_1},F_i^{p_2},...,F_i^{p_k}\}$, where $F_r^{p_k}$ and $F_i^{p_k}$ are pixel-level features from RA and image of point ${p_k}$.
As shown in Figure \ref{fusion}, considering the difference within local distribution and semantics of these two feature spaces, the obtained point-wise features are combined in a hierarchical manner. As low-level and high-level fusions are both efficiently discriminative point-level features, we fuse them at different scales via concatenation after being sequentially processed by a set of shared MLPs. Another key point here is to obtain a per-object global contextual feature which, in principle, reveals the attributes of the same target which shares across domains.
The global point-level feature is obtained via a max-pooling operation of fused features across all candidate points of the same object.
We obtain a set of hybrid point-wise features by concatenating all above mentioned fused features at various scales. These features are fed into a detector that predicts per-point object center locations (see Section \ref{sec:obj_detec}). 

\begin{figure}
[t!]
\centering
\includegraphics[width=0.8\textwidth]{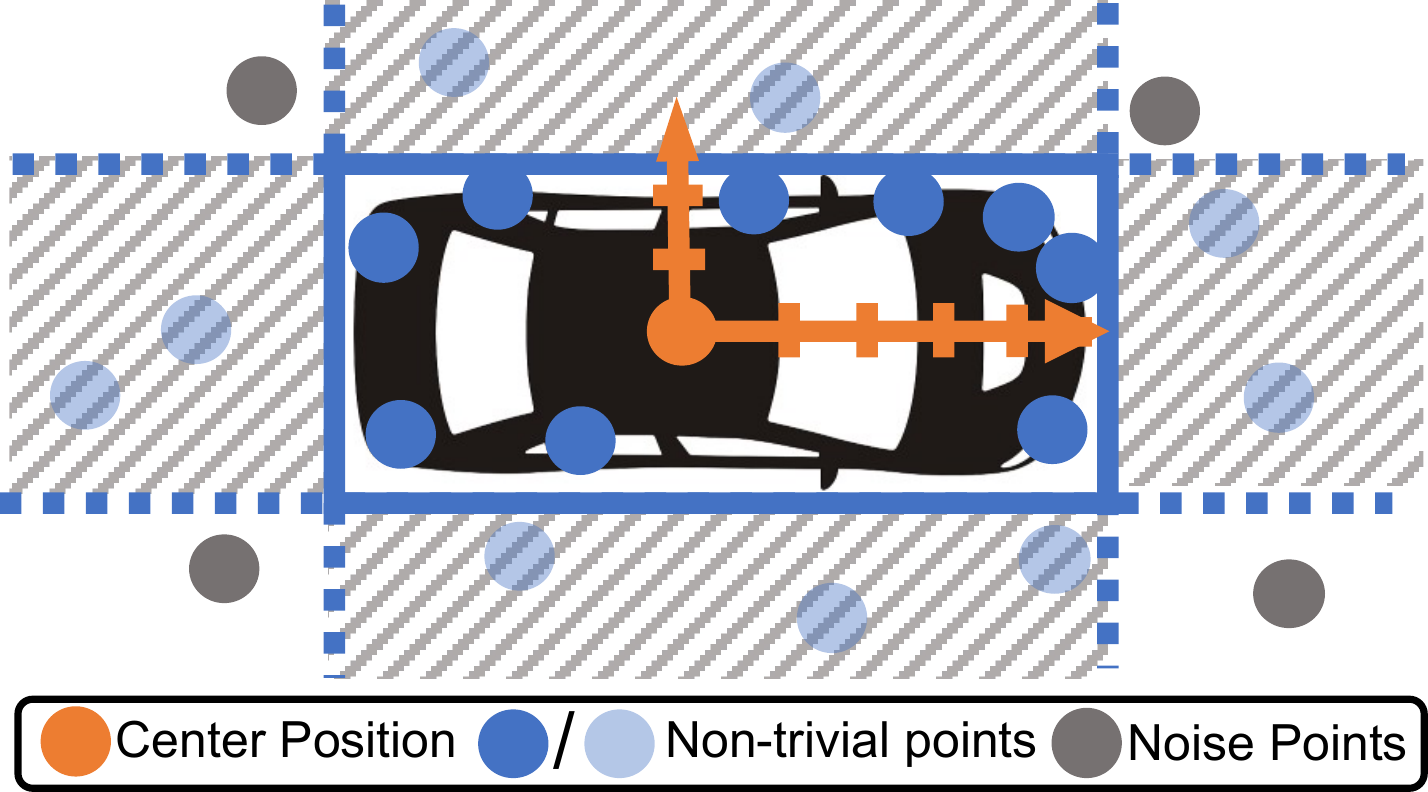}
\caption{Illustration of the local coordinates originated at object center position. Non-trivial points lying within or around the object are kept during training, while noise points are discarded.\label{car}} 
\vspace{-0.5cm}
\end{figure}

\subsection{Object-centric Local Coordinates\label{sec:loc_coord}}

We have experimentally found that directly regressing object locations is not only challenging to achieve purely from extremely sparse point clouds, but also involved with the absolute scale of sensing environments, imaging resolutions and object locations. 
To tackle this problem, we propose to decompose object detection task into a combination of classification and regression sub-tasks at an object-centric local coordinate. 

As shown in Figure~\ref{car}, we establish a new coordinate whose origin is at object center, and $x$, $y$ axes are parallel to range and azimuth axes of RA. For Radar points within 2D bounding boxes, we encode their relative distance to the center position of targets at both axes based on a set of discrete bins at a certain resolution. We further predict a residual offset via regression on top of classification results to reach the final localization prediction.
The motivation comes from the fact that we only focus on the features around the target and this formulation decouples the network prediction from the above-mentioned imaging conditions. In this paper, the relative distance between center labels and Radar points is modeled by discretizing object dimension into 5 bins and 11 bins for azimuth and range, respectively.

It is also worth noting that although Radar points around objects are considered candidates, points which also lie within bounding boxes but are reflected by context near objects also have valuable information. We term such points 'non-trivial' points if their relative distance at \textit{any} axes satisfies our above-mentioned discretization. For example, points reflected by non-object regions within boxes may have a large variation w.r.t. range dimension, but share a large correlation to object at angle dimension. In such cases, we may still include these points as training data and only penalize our network prediction by the deviation at the angle-axis prediction. This investigation also filters radar foreground and background points, eliminating range uncertainty as explained in Section \ref{sec:fusion}.

In the training loop, we use non-trivial points as a data augmentation, which could partly relieve the spatial sparsity of object Radar points.
All other points are regarded as background points or noises and are not involved during network training and inference.
\vspace{-0.25cm}

\subsection{Object Detection and Training Configurations\label{sec:obj_detec}}

As described in Section \ref{sec:loc_coord}, the detection task is divided into two parts, a RA map coarse classification and a refined regression. 
The two-part predictions are trained with a combined loss composed of a Cross-Entropy loss and a Smooth-L1 loss \cite{girshick:2015fast}. Denote the network prediction of classification and regression as $\hat{y}_{cls}^{B\times N\times 16}$ and $\hat{y}_{reg}^{B\times N \times 2}$, the training loss is:
\begin{equation}
    \mathcal{L} = \mathcal{L}_{Cross_Entropy}(y_{cls},\hat{y}_{cls})+\alpha\mathcal{L}_{Smooth-L1}(y_{reg},\hat{y}_{reg}),
\end{equation}
where $\alpha=10$ is a weight balancing parameter.

In the training phase, we use the object's 2D ground truth bounding box to get a precise association. In the test phase, a pertained 2D detector is used to provide object bounding boxes for evaluation. Specifically, we choose to use a pre-trained off-the-shelf YoloX \cite{ge:2021yolox} to obtain 2D bounding boxes on testing images without any fine-tuning on the target datasets. We also show the performance of our method given oracle bounding boxes as a limited case to show the potential upper bound performance of our method.

\section{Experiments\label{sec:exp}}

\subsection{Dataset and Metrics}

\paragraph{Dataset}
We evaluate our model on the RADIal dataset \cite{rebut:2022raw} consisting of RD spectrums and Radar points of a high-definition Radar with corresponding camera observations. Its 91 sequences are divided into hard and easy cases depending on the intensity of Radar perturbation. 
We strictly follow the official splits into training, validation, and test division at a portion of $70\%,15\%,15\%$. Since our proposed point-based architecture requires there are Radar reflection peaks from objects, we remove training candidates where no Radar points are included within object bounding boxes.

\paragraph{Metric}
The evaluation metrics for object detection are Average Precision (AP) and Average Recall (AR), given a validated positive prediction whose Intersection-over-Union (IoU) to the ground truth is greater than $50\%$ \cite{rebut:2022raw}. We also present the absolute Range and Angle error to analyze the prediction accuracy.  

\subsection{Baseline}

\subsubsection{Implement Details}

We implement our image backbone with a pre-trained ResNet-18 \cite{he:2016res18} model.
The color image is of size $960 \times 540$ and we use the semantic features of the last layer as dense image features. The Radar backbone adopts the design of FFT-RadNet \cite{rebut:2022raw} while we further simplify the FPN \cite{lin2017:fpn} model to reduce computational complexity. Due to the high definition nature of used Radar sensor, $\frac{1}{4}$ of native resolution is taken and has been proven to be enough for near-by object discrimination \cite{rebut:2022raw}.   
We train our ROFusion model for $40$ epochs with a batch size of $8$ and $1 \times 10^{-4}$ learning rate with Adam optimizer \cite{kingma:2014adam} on a single NVIDIA Tesla $V100$ GPU. During inference, the bottom Radar points inside objects' 2D bounding boxes are considered as sensor-facing endpoints and are used to generate the heuristic object-centric local coordinates (hLC).

\subsubsection{Results}

\begin{figure}[t!]
\centering
\includegraphics[width=\textwidth]{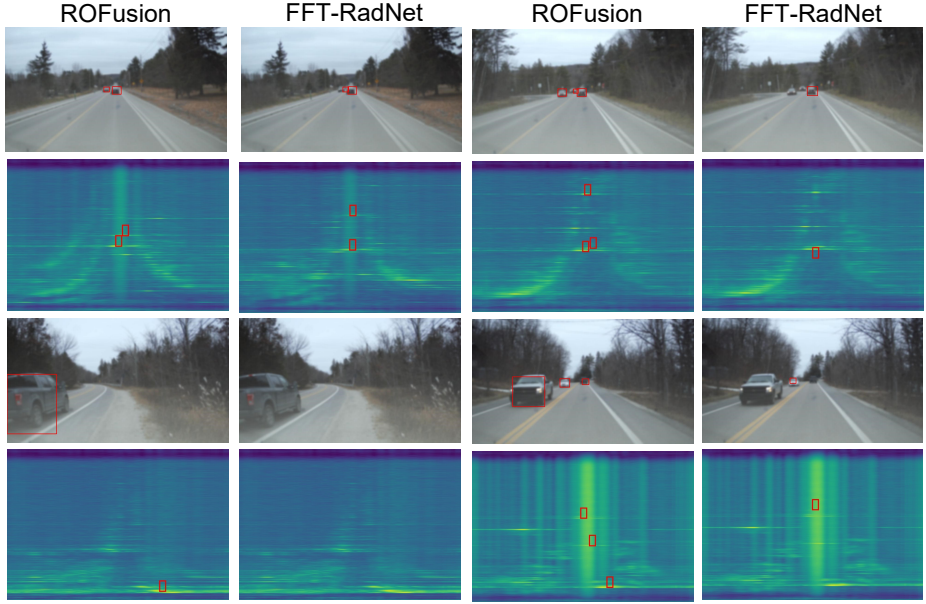}
\vspace{-2em}
\caption{Qualitative results for object detection from Camera (row 1 $\&$ 3) and RA (row 2 $\&$ 4) view. In the RA plots, the detection boxes are presented corresponding to RA dense maps in Euclidean space.} \label{fig:result}
\end{figure}

\begin{table}[t!]
\centering
\caption{Object detection performance on RADIal dataset \cite{rebut:2022raw}. AR ($\%$) is computed with an IoU threshold of $50\%$. R($m$) and A($^\circ$) indicate the mean Range and Angle error. PC, RA, RD and IM mean point clouds, range-azimuth maps, range-doppler maps and images, respectively.}\label{tab:performance}
\vspace{-1em}
\begin{threeparttable}
\scalebox{0.8}{
\begin{tabular}{lcccc|ccc|ccc}
\toprule
\multirow{2.5}{*}{Method} & \multirow{2.5}{*}{Input} &  \multicolumn{3}{c}{Overall} &  \multicolumn{3}{c}{Easy} &  \multicolumn{3}{c}{Hard} \\
\cmidrule{3-11}
 &  &AR(\%) $\uparrow$ & R($m$) $\downarrow$ & A($^\circ$) $\downarrow$ &AR(\%) $\uparrow$ & R($m$) $\downarrow$ & A($^\circ$) $\downarrow$ & AR(\%) $\uparrow$ & R($m$) $\uparrow$ & A($^\circ$) $\downarrow$ \\
\midrule
Pixor \cite{yang:2018pixor} & PC & 32.32  & 0.17 & 0.25 & 28.83 & 0.15 & 0.19 & 38.69 & 0.19 & 0.33\\
Pixor \cite{yang:2018pixor} & RA & 81.68 & \textbf{0.10} & 0.20 & 88.02 & \textbf{0.09} & 0.16 & 70.10 & 0.12 & 0.27\\
FFT-RadNet \cite{rebut:2022raw} & RD & 82.18 & 0.11 & 0.17 & 91.69 &  0.10 & 0.13 & 64.82 & 0.13 & 0.26\\
FFT-RadNet$^*$ \cite{rebut:2022raw} & RD & 82.86 & 0.12 & \textbf{0.11} & 93.12 & 0.11 & \textbf{0.10} & 64.13 & 0.15 & \textbf{0.13} \\ 
\midrule
Ours & IM+RD+PC & \textbf{97.69} & 0.12 & 0.21 & \textbf{97.79} & 0.11 & 0.19 & \textbf{97.52} & \textbf{0.12} & 0.22\\
Ours-hLC & IM+RD+PC & 93.64 & 0.12 & 0.23 & 95.21 & 0.12 & 0.22 & 91.22 & 0.13 & 0.25\\
\bottomrule
\end{tabular}
}
\begin{tablenotes}
\item \small $\bullet$ We denote that FFT-RadNet$*$ \cite{rebut:2022raw} as detector with 0.5 discrimination threshold \\
for a fair comparison, using authors' provided weights.
\end{tablenotes}
\end{threeparttable}
\vspace{-2em}
\end{table}

In Tabel~\ref{tab:performance} we report object detection results of our method compared to leading state-of-the-art methods FFT-RadNet \cite{rebut:2022raw} and baseline method Pixor \cite{yang:2018pixor}. Ground truth bounding boxes and object positions are used to demonstrate the effectiveness of ROFusion. We also evaluate the performance of our method with proposed heuristic local coordinates estimation for practical purposes. Although the radar sparsity causes the worse Angle error, we have observed a clear performance boost despite using sparse point-level features thanks to the optical information and our local coordinate formulation. Our method outperforms \cite{rebut:2022raw} at overall recall rate with a gap of $+14.83\%$. It is worth to highlight that our hybrid point-wise fusion scheme achieves a promising $+27.65\%$ recall boost and a $0.12m$ Range error in the hard cases, overcoming interference problems caused by Radar noise to the dense formulation in \cite{rebut:2022raw}. Qualitative results can be found in Figure \ref{fig:result}.

\begin{table}
    \centering
    \caption{2D Object detection metrics of YOLOX Network \cite{ge:2021yolox} on the test set.}
    \vspace{-1em}
    \label{tab:yolox}
    \setlength{\tabcolsep}{4mm}{
    \scalebox{0.8}{
    \begin{tabular}{lccc|cc|cc}
    \toprule
          \multirow{2.5}{*}{Method} & \multirow{2.5}{*}{Input} &  \multicolumn{2}{c}{Overall} &  \multicolumn{2}{c}{Easy} &  \multicolumn{2}{c}{Hard} \\
    \cmidrule{3-8}
         &  & AP$(\%)$  & AR$(\%)$  & AP$(\%)$ & AR$(\%)$  & AP$(\%)$  & AR$(\%)$  \\
    \midrule
        YOLOX \cite{ge:2021yolox} & IM & 90.48 & 91.03 & 89.79 & 91.86 & 91.77 &89.54\\
        \bottomrule
    \end{tabular}}}
\end{table}

\begin{table}
[t!]
\centering
    \caption{Detection performance on RADIal \cite{rebut:2022raw} given predicted 2D boxes.}\label{tab:test-performance}
    \vspace{-1em}
    \scalebox{0.7}{
    \begin{tabular}{lcccc|cccc|cccc}
    \toprule
    \multirow{2.5}{*}{Method} &  \multicolumn{4}{c}{Overall} &  \multicolumn{4}{c}{Easy} &  \multicolumn{4}{c}{Hard} \\
    \cmidrule{2-13}
    & AP($\%$) $\downarrow$ &AR($\%$) $\uparrow$ & R($m$) $\downarrow$ & A($^\circ$) $\downarrow$ & AP($\%$) $\downarrow$ &AR($\%$) $\uparrow$ & R($m$) $\downarrow$ & A($^\circ$) $\downarrow$ & AP($\%$) $\downarrow$ & AR($\%$) $\uparrow$ & R($m$) $\uparrow$ & A($^\circ$) $\downarrow$\\
    \midrule
    FFT-RadNet$^*$ \cite{rebut:2022raw} & \textbf{97.39} & 82.86 & \textbf{0.12} & \textbf{0.11} & \textbf{98.96} & 93.12 & \textbf{0.11} & \textbf{0.10} & \textbf{93.46} & 64.13 & 0.15 & \textbf{0.13} \\ 
    \midrule
    Ours(YOLOX)-hLC & 91.58 & \textbf{95.15} & 0.13 & 0.21 & 91.03 & \textbf{96.07} & 0.13 & 0.20 & 92.63 & \textbf{93.47} & \textbf{0.13} & 0.23 \\
    \bottomrule
    \end{tabular}}
    \vspace{-2em}
\end{table}

To further demonstrate the practicability of our method, we use network predicted detection results to conduct the evaluation.
We first reveal the quality of the adopted YOLOX \cite{ge:2021yolox} 2D detector in Table \ref{tab:yolox}, with a moderate performance drop in terms of optical detection accuracy,  it is expected that the imperfect 2D detection results would affect the filtering process of our pipeline.
Table ~\ref{tab:test-performance} compares the $AP$ and $AR$ metrics with machine-generated bounding boxes in both easy and hard cases as well.
While the AP metric lack behind baseline model due to the quality of network inferred 2D bounding boxes, our method with YOLOX \cite{ge:2021yolox} 2D detector still achieves a higher AR metric for both overall and especially difficult cases. The AR performance gain comes from the 2D bounding boxes association and heuristic local coordinates. The Range error of our method also outperforms FFT-RadNet$*$ \cite{rebut:2022raw} in difficult cases. These results show that our local coordinates successfully extract the range information for sparse Radar points even though the prior 2D detection is less accurate.

\subsection{Ablation}

In this section, we conduct ablative experiments to validate the key components of our method: local coordinate (LC) formulation and image fusion module (IM).

As shown in Table \ref{ablation}, we have shown two variants of ROFusion: Ours (w/o LC) is the variant where we remove the local coordinate formulation in the training stage, but conduct the two classification and regression sub-tasks in the original RA maps; we also remove the image level feature fusion module out of training process. From the statistics, we conclude that the local coordinate formulation is significant in enabling accurate learning from spare Radar point clouds. The integration of image features gives further performance boost upon competing performance. 
It is also worth noting that in addition to the image feature, the prominent prior information introduced by optical detection is another key factor supporting the overall learning process from sparse Radar point clouds.

\vspace{-0.5em}
\section{Conclusion}
\vspace{-0.5em}

\begin{table}
[ht]
\centering
\caption{Ablation study on the key components of ROFusion.}\label{ablation}
\vspace{-1em}
\scalebox{0.8}{
\begin{tabular}{lcccc|ccc|ccc}
\toprule
\multirow{2.5}{*}{Method} & \multirow{2.5}{*}{Input} &  \multicolumn{3}{c}{Overall} &  \multicolumn{3}{c}{Easy} &  \multicolumn{3}{c}{Hard} \\
\cmidrule{3-11}
 &  &AR($\%$) $\uparrow$ & R($m$) $\downarrow$ & A($^\circ$) $\downarrow$ &AR($\%$) $\uparrow$ & R($m$) $\downarrow$ & A($^\circ$) $\downarrow$ & AR($\%$) $\uparrow$ & R($m$) $\downarrow$ & A($^\circ$) $\downarrow$ \\
 \midrule
 Ours (w/o LC) & RD+PC & 27.4 & 0.22 & 0.46 & 32.94 & 0.24 & 0.40 & 15.78 & 0.20 & 0.57\\
Ours & RD+PC & 96.58 & \textbf{0.09} & 0.22 & 96.31 & \textbf{0.08} & 0.22 & 96.85 & \textbf{0.09} & 0.24\\
Ours & IM+RD+PC & \textbf{97.69} & 0.12 & \textbf{0.21} & \textbf{97.79} & 0.11 & \textbf{0.19} & \textbf{97.52} & 0.12 & \textbf{0.22}\\
\bottomrule
\end{tabular}}
\vspace{-2em}
\end{table}

In this paper, we present ROFusion, a novel point-wise Radar-Optical fusion network for object detection. 
We have demonstrated that our method could effectively exploit camera semantics to enhance Radar detection. With hybrid point fusion and local coordinate formulation, ROFusion achieves state-of-the-art performance on the public RADIal dataset \cite{rebut:2022raw}, showing the potential capability for multi-sensor fusion. 
However, our method still relied on the quality of 2D object detection as prior information to filter potential object Radar points. In addition, considering the difference in imaging mechanism, more in-depth analysis of camera-Radar fusion stratify at the feature level is worth investigating, possibly aided by a powerful Transformer backbone using attention mechanism. 
This can be an exciting venue for our future work. 
\vspace{-0.5em}

\section*{Acknowledgements}
\vspace{-0.5em}

This work was partially supported by the National Key Research and
Development Program of China (Grant No. 2021YFB3100800), the National Natural Science Foundation of China (Grant No. 61921001, 62201603) and Research Program of National University of Defense Technology (Grant No. ZK22-04).

\bibliography{reference}

\vspace{-1em}

\begin{thebibliography}{10}

\bibitem{cesic:2016radar}
Josip {\'C}esi{\'c}, Ivan Markovi{\'c}, Igor Cvi{\v{s}}i{\'c}, and Ivan
  Petrovi{\'c}.
\newblock Radar and stereo vision fusion for multitarget tracking on the
  special euclidean group.
\newblock {\em Robotics and Autonomous Systems}, 83:338--348, 2016.

\bibitem{chen：2017multilidar}
Xiaozhi Chen, Huimin Ma, Ji~Wan, Bo~Li, and Tian Xia.
\newblock Multi-view 3d object detection network for autonomous driving.
\newblock In {\em {Proceedings of the {IEEE} Conference on Computer Vision and
  Pattern Recognition ({CVPR})}}, pages 1907--1915, 2017.

\bibitem{dreher:2020radar}
Maria Dreher, Eme{\c{c}} Er{\c{c}}elik, Timo B{\"a}nziger, and Alois Knol.
\newblock Radar-based 2d car detection using deep neural networks.
\newblock In {\em {Proceedings of the International {IEEE} Conference on
  Intelligent Transportation Systems ({ITSC})}}, pages 1--8, 2020.

\bibitem{gaisser:2017road}
Floris Gaisser and Pieter~P Jonker.
\newblock Road user detection with convolutional neural networks: An
  application to the autonomous shuttle wepod.
\newblock In {\em {Journal of Machine Vision and Applications}}, pages
  101--104. IEEE, 2017.

\bibitem{ge:2021yolox}
Zheng Ge, Songtao Liu, Feng Wang, Zeming Li, and Jian Sun.
\newblock Yolox: Exceeding yolo series in 2021.
\newblock {\em arXiv preprint arXiv:2107.08430}, 2021.

\bibitem{girshick:2015fast}
Ross Girshick.
\newblock Fast r-cnn.
\newblock In {\em {Proceedings of the International Conference on Computer
  Vision ({ICCV})}}, pages 1440--1448, 2015.

\bibitem{guo:2018pedestrian}
Xiao-peng Guo, Jin-song Du, Jie Gao, and Wei Wang.
\newblock Pedestrian detection based on fusion of millimeter wave radar and
  vision.
\newblock In {\em {International Conference on Artificial Intelligence and
  Pattern Recognition}}, pages 38--42, 2018.

\bibitem{he:2016res18}
Kaiming He, Xiangyu Zhang, Shaoqing Ren, and Jian Sun.
\newblock Deep residual learning for image recognition.
\newblock In {\em {Proceedings of the {IEEE} Conference on Computer Vision and
  Pattern Recognition ({CVPR})}}, pages 770--778, 2016.

\bibitem{hwang:2022cramnet}
Jyh-Jing Hwang, Henrik Kretzschmar, Joshua Manela, Sean Rafferty, Nicholas
  Armstrong-Crews, Tiffany Chen, and Dragomir Anguelov.
\newblock Cramnet: Camera-radar fusion with ray-constrained cross-attention for
  robust 3d object detection.
\newblock In {\em {Proceedings of the European Conference on Computer Vision
  ({ECCV})}}, pages 388--405. Springer, 2022.

\bibitem{kim:2022craft}
Youngseok Kim, Sanmin Kim, Jun~Won Choi, and Dongsuk Kum.
\newblock Craft: Camera-radar 3d object detection with spatio-contextual fusion
  transformer.
\newblock {\em arXiv preprint arXiv:2209.06535}, 2022.

\bibitem{kingma:2014adam}
Diederik~P Kingma and Jimmy Ba.
\newblock Adam: A method for stochastic optimization.
\newblock {\em arXiv preprint arXiv:1412.6980}, 2014.

\bibitem{lin2017:fpn}
Tsung-Yi Lin, Piotr Doll{\'a}r, Ross Girshick, Kaiming He, Bharath Hariharan,
  and Serge Belongie.
\newblock Feature pyramid networks for object detection.
\newblock In {\em {Proceedings of the {IEEE} Conference on Computer Vision and
  Pattern Recognition ({CVPR})}}, pages 2117--2125, 2017.

\bibitem{liu:2022deep}
Jianan Liu, Weiyi Xiong, Liping Bai, Yuxuan Xia, Tao Huang, Wanli Ouyang, and
  Bing Zhu.
\newblock Deep instance segmentation with automotive radar detection points.
\newblock {\em {{IEEE} Transactions on Intelligent Vehicles}}, 2022.

\bibitem{nabati:2021centerfusion}
Ramin Nabati and Hairong Qi.
\newblock Centerfusion: Center-based radar and camera fusion for 3d object
  detection.
\newblock In {\em {Proceedings of the {IEEE} Workshop on Applications of
  Computer Vision ({WACV})}}, pages 1527--1536, 2021.

\bibitem{qi:2017pointnet}
Charles~R Qi, Hao Su, Kaichun Mo, and Leonidas~J Guibas.
\newblock Pointnet: Deep learning on point sets for 3d classification and
  segmentation.
\newblock In {\em {Proceedings of the {IEEE} Conference on Computer Vision and
  Pattern Recognition ({CVPR})}}, pages 652--660, 2017.

\bibitem{qi:2017pointnet++}
Charles~Ruizhongtai Qi, Li~Yi, Hao Su, and Leonidas~J Guibas.
\newblock Pointnet++: Deep hierarchical feature learning on point sets in a
  metric space.
\newblock {\em {Neural Information Processing Systems ({NIPS})}}, 30, 2017.

\bibitem{rebut:2022raw}
Julien Rebut, Arthur Ouaknine, Waqas Malik, and Patrick P{\'e}rez.
\newblock Raw high-definition radar for multi-task learning.
\newblock In {\em Proceedings of the IEEE/CVF Conference on Computer Vision and
  Pattern Recognition}, pages 17021--17030, 2022.

\bibitem{russakovsky:2015imagenet}
Olga Russakovsky, Jia Deng, Hao Su, Jonathan Krause, Sanjeev Satheesh, Sean Ma,
  Zhiheng Huang, Andrej Karpathy, Aditya Khosla, Michael Bernstein, et~al.
\newblock Imagenet large scale visual recognition challenge.
\newblock {\em {International Journal of Computer Vision ({IJCV})}},
  115:211--252, 2015.

\bibitem{scheiner:2021object}
Nicolas Scheiner, Florian Kraus, Nils Appenrodt, J{\"u}rgen Dickmann, and
  Bernhard Sick.
\newblock Object detection for automotive radar point clouds--a comparison.
\newblock {\em AI Perspectives}, 3(1):1--23, 2021.

\bibitem{vaswani:2017attention}
Ashish Vaswani, Noam Shazeer, Niki Parmar, Jakob Uszkoreit, Llion Jones,
  Aidan~N Gomez, {\L}ukasz Kaiser, and Illia Polosukhin.
\newblock Attention is all you need.
\newblock {\em Advances in neural information processing systems}, 30, 2017.

\bibitem{vora:2020pointpainting}
Sourabh Vora, Alex~H Lang, Bassam Helou, and Oscar Beijbom.
\newblock Pointpainting: Sequential fusion for 3d object detection.
\newblock In {\em {Proceedings of the {IEEE} Conference on Computer Vision and
  Pattern Recognition ({CVPR})}}, pages 4604--4612, 2020.

\bibitem{wang2019:densefusion}
Chen Wang, Danfei Xu, Yuke Zhu, Roberto Mart{\'\i}n-Mart{\'\i}n, Cewu Lu,
  Li~Fei-Fei, and Silvio Savarese.
\newblock Densefusion: 6d object pose estimation by iterative dense fusion.
\newblock In {\em Proceedings of the IEEE/CVF conference on computer vision and
  pattern recognition}, pages 3343--3352, 2019.

\bibitem{yang:2018pixor}
Bin Yang, Wenjie Luo, and Raquel Urtasun.
\newblock Pixor: Real-time 3d object detection from point clouds.
\newblock In {\em {Proceedings of the {IEEE} Conference on Computer Vision and
  Pattern Recognition ({CVPR})}}, pages 7652--7660, 2018.

\bibitem{zhong：2018camera}
Ziguo Zhong, Stanley Liu, Manu Mathew, and Aish Dubey.
\newblock Camera radar fusion for increased reliability in adas applications.
\newblock {\em Electronic Imaging}, 2018(17):258--1, 2018.

\end{thebibliography}

\end{document}